\documentclass{article}
\usepackage{ijcai25}

\usepackage{times}
\usepackage{soul}
\usepackage{url}
\usepackage[hidelinks]{hyperref}
\usepackage[utf8]{inputenc}
\usepackage[small]{caption}
\usepackage{graphicx}
\usepackage{amsmath}
\usepackage{amsthm}
\usepackage{booktabs}
\usepackage{algorithm}
\usepackage{algorithmic}
\usepackage[switch]{lineno}
\usepackage{xcolor}
\usepackage{listings}
\title{Lifelong Reinforcement Learning with Similarity-Driven Weighting by Large Models}

\author{
Zhiyi Huang$^1$
\and
Xiaohan Shan$^1$\and
Jianmin Li$^1$\\
\affiliations
$^1$Qiyuan Lab\\
\emails
\{huangzhiyi,shanxiaohan,lijianmin\}@qiyuanlab.com
}

\begin{document}

\maketitle

\begin{abstract}
    Lifelong Reinforcement Learning (LRL) holds significant potential for addressing sequential tasks, but it still faces considerable challenges. A key difficulty lies in effectively preventing catastrophic forgetting and facilitating knowledge transfer while maintaining reliable decision-making performance across subsequent tasks in dynamic environments. To tackle this, we propose a novel framework, SDW (Similarity-Driven Weighting Framework), which leverages large-language-model-generated dynamic functions to precisely control the training process. The core of SDW lies in two functions pre-generated by large models: the task similarity function and the weight computation function. The task similarity function extracts multidimensional features from task descriptions to quantify the similarities and differences between tasks in terms of states, actions, and rewards. The weight computation function dynamically generates critical training parameters based on the similarity information, including the proportion of old task data stored in the Replay Buffer and the strategy consistency weight in the loss function, enabling an adaptive balance between learning new tasks and transferring knowledge from previous tasks. By generating function code offline prior to training, rather than relying on large-model inference during the training process, the SDW framework reduces computational overhead while maintaining efficiency in sequential task scenarios. Experimental results on Atari and MiniHack sequential tasks demonstrate that SDW significantly outperforms existing lifelong reinforcement learning methods.
\end{abstract}

\section{Introduction}

Reinforcement Learning (RL) has achieved remarkable success in solving single tasks, such as game control, robotic manipulation, and decision optimization. 
However, in real-world scenarios, intelligent systems often encounter a sequence of tasks that require continuous learning and adaptation while retaining previously acquired knowledge. To address these challenges, researchers have proposed a class of algorithms known as Lifelong Reinforcement Learning (LRL), which aim to enable agents to solve sequential decision-making problems in dynamic environments. LRL has gained increasing attention in recent research due to its importance in applications such as autonomous robotics\cite{liu2019lifelong}, games\cite{sur2022system}, and adaptive control\cite{lu2019adaptive}, where agents must not only acquire new skills but also maintain strong performance on previously learned tasks. However, designing effective LRL algorithms that balance learning new tasks and retaining past knowledge remains a significant challenge.

\begin{figure}[htbp]
    \centering
    \includegraphics[width=\linewidth]{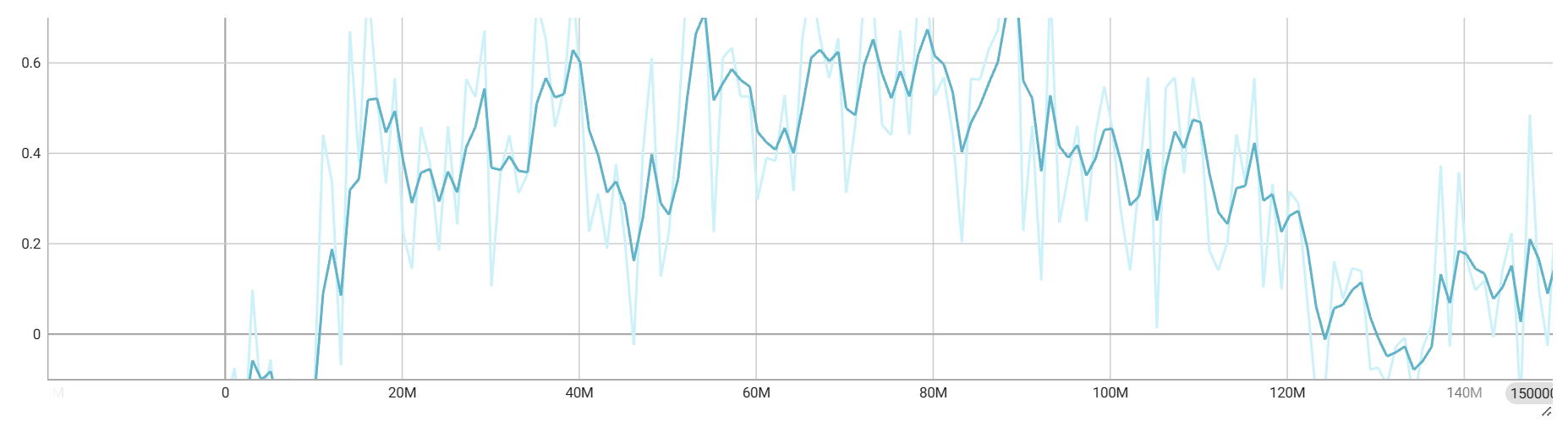}
    \caption{Performance of the CLEAR algorithm on the MiniHack benchmark. The x-axis represents training steps, and the y-axis represents the reward. A new task is introduced every 10 million steps.}
    \label{fig:clear_minihack}
\end{figure}

The practical challenges of existing LRL methods, such as CLEAR\cite{rolnick2019experience} and Elastic Weight Consolidation (EWC) \cite{kirkpatrick2017overcoming}, become evident when applied to tasks with significant differences. 
We validated the CLEAR algorithm in the MiniHack environment, as illustrated in Figure \ref{fig:clear_minihack}. The results show a noticeable decline in performance on later tasks. This behavior can be attributed to its strategy of combining data from earlier tasks with new task data during training. When the gap between tasks is large, this approach struggles to reconcile conflicting task objectives, making it difficult for the algorithm to effectively optimize for the new task. The shared training process introduces interference, as the agent is forced to balance learning objectives that are not well-aligned across tasks.

\begin{figure}[htbp]
    \centering
    \includegraphics[width=\linewidth]{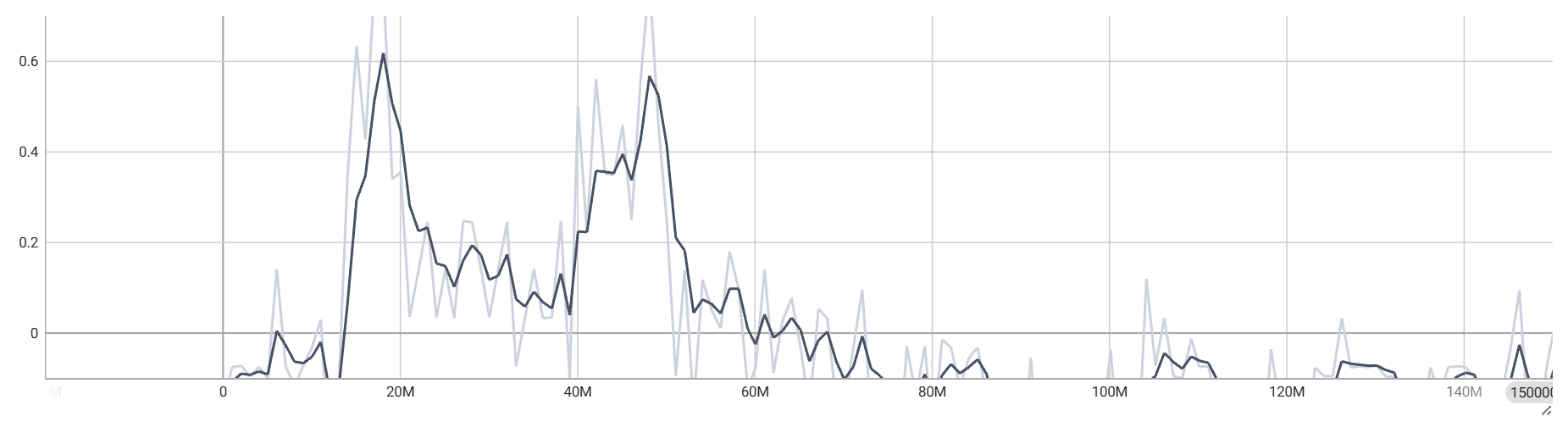}
    \caption{Performance of the EWC algorithm on the MiniHack benchmark. The x-axis represents training steps, and the y-axis represents the reward. A new task is introduced every 10 million steps.}
    \label{fig:ewc_minihack}
\end{figure}

A similar phenomenon is observed with EWC, as shown in Figure \ref{fig:ewc_minihack}. EWC attempts to retain performance on earlier tasks by constraining parameter updates during new task training. However, this rigid preservation of earlier knowledge does not account for task relationships and instead imposes strict limitations on the decision-making space for subsequent tasks. When earlier tasks are fundamentally different from later tasks, these constraints hinder the agent’s ability to fully adapt to the new task, leading to suboptimal performance.

These observations highlight a fundamental limitation of existing methods: both CLEAR and EWC fail to model and leverage task relationships effectively. By treating tasks as independent and enforcing rigid constraints or replay mechanisms, they are unable to dynamically adapt to the evolving nature of task sequences. This underscores the need for more advanced approaches that can explicitly model inter-task dependencies, adaptively balance knowledge retention and new task learning, and address the unique challenges posed by diverse and complex task sequences.

Recently, large language models (LLMs) have achieved groundbreaking success in natural language processing\cite{achiam2023gpt} and other fields such as law\cite{cheong2024not}, medical\cite{wang2024jmlr} and communication\cite{kan2024mobile}, demonstrating their ability to extract and utilize complex relationships in data\cite{hong2024data}. These capabilities make LLMs a promising tool for addressing the challenges of LRL, particularly in analyzing task relationships and guiding multi-task learning sequences.

Inspired by these advancements, we propose a novel LRL framework that leverages LLMs to improve task performance, mitigate forgetting, and enhance transferability. The framework introduces two key functions: one to quantitatively analyze task relationships and measure their relevance, and another to dynamically map task relevance to critical training parameters, such as loss calculation and buffer updates. This approach enables agents to adaptively control their training process, improving task retention while accelerating the learning of new tasks.

Unlike previous methods that require deep involvement of LLMs, our framework uses LLMs only during the function generation stage, ensuring lightweight integration. By effectively capturing task relationships and dynamically adjusting the training process, our framework significantly outperforms existing methods in task performance, forgetting mitigation, and knowledge transfer. Furthermore, it offers a novel perspective for integrating large models into LRL, advancing the development of AI systems capable of lifelong learning.

The main contributions of this work are as follows:

\begin{itemize}
    \item We propose a novel lifelong reinforcement learning framework, SDW, which leverages the task relationship reasoning capabilities of LLMs to dynamically optimize multi-task learning processes. Compared to traditional methods, SDW significantly improves the efficiency of learning new tasks while effectively mitigating forgetting, providing a new solution for lifelong learning.
    \item We guide the large language model to generate two key functions: a task similarity calculation function, which quantitatively evaluates the relationships between tasks based on their descriptions, state distributions, and reward structures, and a weight calculation function, which determines the optimal training weights for balancing new and old tasks. By integrating these functions into the training process, we achieve a more reasonable balance between learning new tasks and the retention of old tasks.
    \item We implement a lightweight LLM integration strategy that utilizes LLMs only during the pretraining phase for task relationship reasoning, making the training process fully independent of LLMs. This design allows us to significantly reduce computational overhead while avoiding reliance on high-performance LLMs, ensuring that our framework remains efficient and scalable when handling large-scale task sequences.
    \item Our experimental results on MiniHack and Atari demonstrate that SDW significantly outperforms existing algorithms in mitigating catastrophic forgetting and improving the learning of new tasks, highlighting its effectiveness in lifelong reinforcement learning scenarios.
\end{itemize}

\section{Related Work}
\subsection{Lilfelong Reinforcement Learning}

Lifelong reinforcement learning aims to endow agents with the ability to accumulate knowledge over sequential tasks while mitigating catastrophic forgetting. Prior studies in LRL can be broadly categorized into three approaches:

\subsubsection{Regularization-based Knowledge Retention}

These methods use regularization terms to preserve knowledge and reduce interference with earlier tasks. Notable examples include Elastic Weight Consolidation (EWC) \cite{kirkpatrick2017overcoming}and Online EWC\cite{huszar2018note}, which mitigate forgetting by allowing suboptimal solutions for new tasks. While effective, these methods often face limitations in scalability and adaptability to highly dynamic task environments.

\subsubsection{Experience Replay-based Methods}
Experience replay methods store past task experiences and replay them during training, enabling agents to avoid catastrophic forgetting while improving learning performance on new tasks. Algorithms such as CLEAR\cite{rolnick2019experience} and PnC\cite{schwarz2018progress} fall into this category. Although these methods improve task performance to some extent, they struggle in scenarios with large task spaces due to their lack of task specificity and inefficiencies in prioritizing relevant experiences.
\subsubsection{Network Structure-based Methods}
These approaches dynamically adjust the network architecture, such as freezing specific parameters or adding new modules, allowing agents to flexibly adapt to new tasks. An example is the SANE\cite{powers2022self}, which effectively isolates task-specific knowledge. However, these methods tend to increase network complexity significantly, limiting their practicality in large-scale task deployments.
\subsection{Advances in LLMs and Lifelong Learning}
To overcome the limitations of traditional LRL methods in dynamic task environments, researchers have begun exploring the integration of LLMs, which possess powerful reasoning and knowledge transfer capabilities, into lifelong reinforcement learning frameworks. Meanwhile, studies on the lifelong learning capabilities of LLMs have also emerged. 

For instance, \cite{sun2019lamol} proposed LAMOL, a simple yet effective method for lifelong language learning (LLL) based on language modeling.LAMOL addresses catastrophic forgetting by solving current tasks while generating pseudo-samples of previous tasks for joint training. However, it is limited to language tasks and does not generalize well to other domains, such as robotics or games.

Additionally, research on the combination of LLMs and RL has made progress as well. Research also explores combining LLMs with reinforcement learning, focusing on task exploration and goal generation.  \cite{pourcel2024autotelic} proposed an autotelic agent framework that combines LLMs with goal-conditioned RL to address open-ended goal generation and learning. The LLM estimates goal difficulty and learnability based on context, generating reward function codes and goal descriptions to optimize the goal generation and learning process. This approach outperforms traditional methods in terms of goal generation efficiency and adaptation speed. However, it does not address catastrophic forgetting in multi-task scenarios, which remains a critical limitation in LRL.
\subsection{Combining LLMs with LRL}
While prior works have explored the potential of LLMs in lifelong learning and reinforcement learning separately, recent studies focus on integrating LLMs into lifelong reinforcement learning frameworks to address the unique challenges of dynamic multi-task environments.
The integration of large language models with lifelong reinforcement learning has emerged as a promising direction, leveraging LLMs' reasoning capabilities and vast prior knowledge to enhance the adaptability and scalability of LRL frameworks. Current research primarily falls into two main categories:
\subsubsection{Using LLMs as Planners to Build Skill Graphs}
This approach leverages LLMs to construct skill graphs, enabling agents to identify actions for complex tasks. \cite{mao2024framework} proposed a framework where LLMs are used to select skills for agents and determine whether new skills are necessary. Similarly, \cite{yuan2023skill}leveraged LLMs' prior knowledge to discover relationships between skills, constructing skill graphs to explore open-world environments such as Minecraft.

Li et al. \cite{li2024empowering} introduced the LEAGUE framework, which integrates LLMs with Task and Motion Planning and Deep Reinforcement Learning. LLMs decompose tasks, create skill operators, and generate dense rewards to accelerate policy learning while maintaining a skill library for new problem scenarios. Tziafas and Kasaei \cite{tziafas2024lifelong} proposed an approach to continuously expand a robotic skill library. Their approach alternates between a ‘wake’ phase, where an LLM generates and validates tasks in a simulator, and a ‘sleep’ phase, where the LLM abstracts experiences into new skills and optimizes knowledge via memory replay.
\subsubsection{Using LLMs as Semantic Generators for Observations}
Another line of research treats LLMs as an environment-like entity, where task-relevant language generated by the LLM is encoded and embedded into the state space to facilitate knowledge transfer and task learning. 

\cite{meng2024preserving} proposed a framework that embeds semantic information generated by LLMs into task states, enabling effective knowledge transfer. The system uses the SAC algorithm to learn task actions by encoding task descriptions as language embeddings, which are incorporated into the state space during training.

\cite{chenlearning} utilized pretrained LLMs (e.g., GPT-3.5) to generate task hints and content from textual descriptions. These task-related embeddings are combined with observation information, input into a transformer model, and used to generate language-guided rewards. This approach helps policies adapt better to tasks.

Although research in this area is still in its early stages, these methods show great potential for enhancing task understanding and enabling efficient knowledge transfer in lifelong learning scenarios. However, expanding the state space increases complexity, requiring more iterations and computational resources. Furthermore, continuous interaction with large models during training significantly increases computational demand, leading to longer training times and requiring high-performance hardware to process the expanded state space efficiently.
\begin{figure*}[htbp]
\centering 
\includegraphics[width=\linewidth]{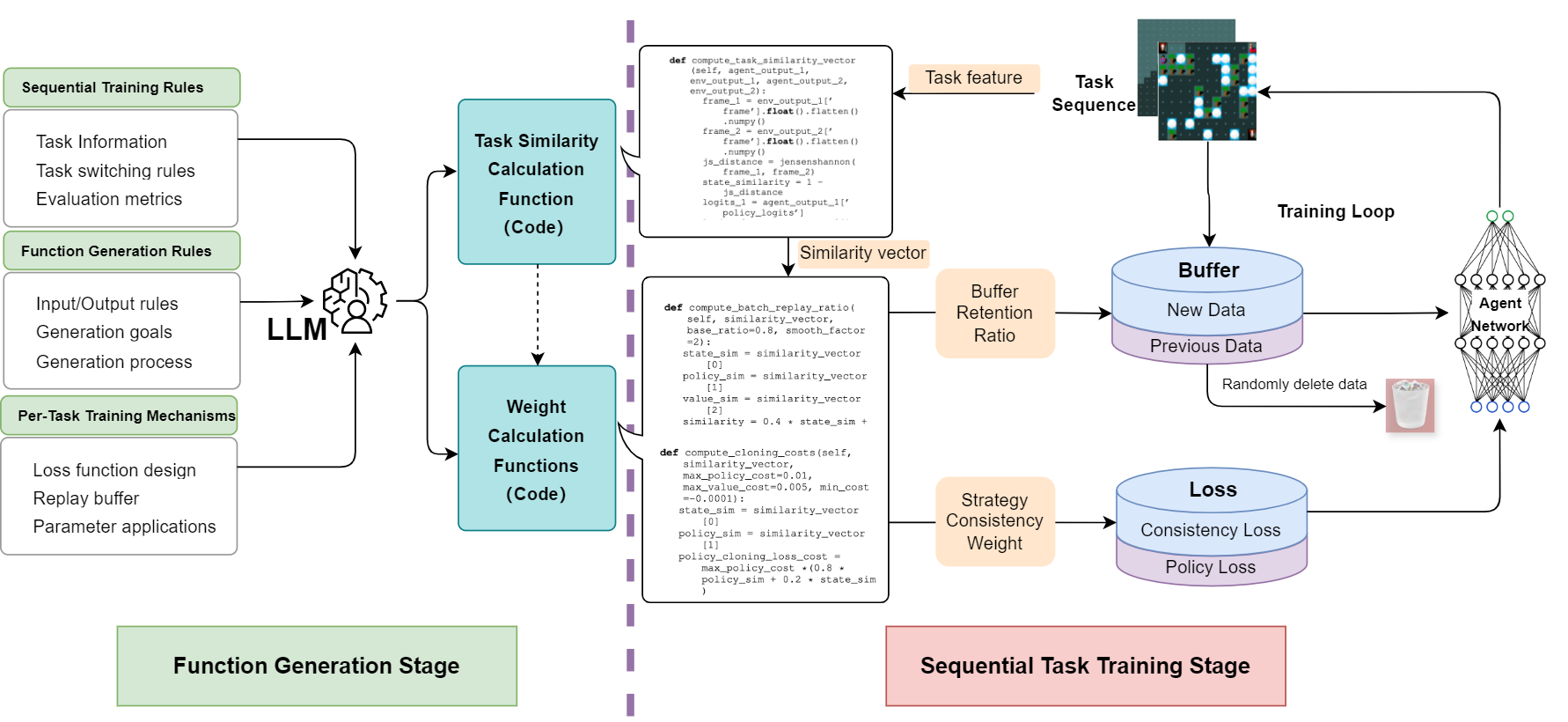}
\caption{The Workflow of SDW.} 
\label{fig_framework} 
\end{figure*}
\section{Method}
We propose a Similarity-Driven Weighting (SDW) framework, a lifelong reinforcement learning algorithm designed to address the challenge of catastrophic forgetting in sequential task learning. The core idea of SDW is to adaptively determine key parameter settings in the continual training process based on the relationships between tasks. Instead of relying on fixed or manually tuned parameters, SDW leverages the reasoning capabilities of large models to analyze task similarities and generate two essential functions: a task similarity calculation function and a weight calculation function. These functions dynamically quantify the extent of similarity between tasks and adjust the retention of knowledge—both in terms of data and policies—from previously learned tasks. By tailoring the degree of knowledge transfer and retention to the specific relationships between tasks, SDW ensures a more flexible and efficient learning process across task sequences, significantly improving performance in lifelong reinforcement learning scenarios.
We provide more in-depth insights into the design of SDW in the Appendix.

\subsection{Framework Workflow}

The SDW framework operates in two main stages: the function generation stage and the sequential task training stage, as illustrated in Figure \ref{fig_framework}.
During the function generation stage, SDW leverages LLMs to generate two key functions: the Task Similarity Calculation Function and the Weight Calculation Function. The prompts for the LLM are constructed using task information (e.g., task characteristics, continual learning process, replay buffer usage) and design requirements, which specify the desired outputs of these functions. This information provides essential contextual knowledge about the relationships between tasks in a lifelong learning setting. The Task Similarity Calculation Function computes a similarity vector between consecutive tasks based on their environmental encodings, which represents the relationship between the current task and the immediately preceding task. This vector captures various aspects of task similarity and serves as the foundation for subsequent adjustments. The Weight Calculation Function uses the similarity vector to dynamically adjust key parameters during the training process, ensuring that the framework effectively balances knowledge retention and transfer across tasks.

In the sequential task training stage, the framework applies the functions generated by the LLM to process and optimize the continual learning process. Specifically, the similarity vector computed by the Task Similarity Calculation Function informs the Weight Calculation Function to adjust sampling weights for the replay buffer and key weights in the loss function. These adjustments enable the framework to incrementally train the model on new tasks while mitigating catastrophic forgetting.

What needs to be emphasized is that SDW confines the use of LLMs to a one-time inference during the function generation stage, after which the generated functions are applied throughout the sequential task training phase to compute task-specific weights and adjustments without requiring further LLM inference. This design avoids embedding LLM inference into the training process, significantly reducing computational overhead and ensuring efficient and scalable training. Unlike previous works (e.g., \cite{meng2024preserving}, \cite{chenlearning}) that rely on frequent LLM inference during training, SDW decouples LLM usage from the training loop, allowing us to leverage high-parameter LLMs without compromising efficiency.

\subsection{Function Generation}
In the function generation phase, SDW leverages LLMs to automatically generate two core functions: the task similarity function and the weight computation function. These two functions are used to quantify the multidimensional similarity between tasks and to dynamically generate training parameters based on the computed similarity. 
The input for the task similarity function is generated from the descriptions of the tasks. Task descriptions may include information such as objectives, state space, action space, and reward functions. Based on these descriptions, the LLM extracts task features and generates a feature representation that characterizes the task.

The task similarity function outputs a multidimensional similarity vector, denoted as 
$\mathbf{S} = [S_{\text{state}}, S_{\text{action}}, S_{\text{reward}}, \dots]$,
which quantifies task similarity across multiple features, such as states, actions, and rewards. Each dimension in $\mathbf{S}$ lies in the range $[0, 1]$, where values closer to 1 indicate greater similarity for the respective feature, and values closer to 0 indicate significant differences. The function is formally defined as:
\[
\mathcal{F}_{\text{similarity}}: (\mathbf{d}_{\text{current}}, \mathbf{d}_{\text{previous}}) \rightarrow \mathbf{S},
\]
where $\mathbf{d}_{\text{current}}$ and $\mathbf{d}_{\text{previous}}$ are feature vectors extracted from the descriptions of the current and previous tasks. The logic for feature extraction and similarity computation is automatically generated by the LLM, ensuring that the function captures nuanced, multidimensional relationships between tasks. This approach enables a more comprehensive understanding of the differences and similarities across tasks, which is critical for effective continual learning.

The weight computation function uses the similarity vector $\mathbf{S}$ to dynamically determine two key training parameters: the weight $w_{\text{buffer}}$ for adjusting the proportion of old task data in the replay buffer, and the weight $\lambda_{\text{consistency}}$ for scaling the strategy distance loss in the total loss function. Formally, the function is defined as:
\[
\mathcal{F}_{\text{weight}}: \mathbf{S} \rightarrow (w_{\text{buffer}}, \lambda_{\text{consistency}}),
\]
where both $w_{\text{buffer}}$ and $\lambda_{\text{consistency}}$ lie in the range $[0, 1]$. Higher similarity values in $\mathbf{S}$ lead to a larger $w_{\text{buffer}}$, preserving more old task data in the replay buffer to emphasize knowledge retention. Similarly, higher similarity increases $\lambda_{\text{consistency}}$, prioritizing consistency with the previous task’s strategy by assigning greater importance to the strategy distance loss.

By leveraging the multidimensional information in $\mathbf{S}$, the LLM-generated weight computation function flexibly adjusts these parameters based on the specific relationships between tasks. This dynamic adjustment ensures a balance between knowledge retention and adaptation, enabling the framework to optimize training efficiency while mitigating catastrophic forgetting. The integration of these functions forms the foundation of the SDW framework’s adaptive approach to continual learning in a principled and adaptive manner. 


\subsection{Sequential Task Training}
In sequential task training, the SDW framework dynamically adjusts key training parameters through the synergy of the task similarity function and the weight computation function, balancing the utilization of historical knowledge with the learning requirements of the new task. Specifically, the task similarity function generates a multidimensional similarity vector $\mathbf{S}$ based on features like states, actions, and rewards, which provides a precise characterization of the relationship between the new and old tasks. Using $\mathbf{S}$ as input, the weight computation function produces two critical parameters: $w_{\text{buffer}}$, which controls the proportion of old task data in the replay buffer, and $\lambda_{\text{consistency}}$, which regulates the weight of the strategy consistency loss in the total loss function. 

\textbf{Replay Buffer Management.} The dynamic management of the replay buffer is governed by $w_{\text{buffer}}$, which determines the balance between old and new task data in the buffer. A higher $w_{\text{buffer}}$ retains more old task data, strengthening the utilization of historical knowledge, while a lower $w_{\text{buffer}}$ reduces the proportion of old task data, minimizing interference with new task learning. To maintain this balance, SDW employs a dynamic insertion probability $P_{\text{insert}}$ to control the entry of new task data into the buffer:
$P_{\text{insert}} = P_{\text{base}} + \lambda \cdot \left(1 - \frac{w_{\text{buffer}}}{p_{\text{old}}}\right)$,
where $P_{\text{base}}$ is a base insertion probability that ensures new task data always has some opportunity to enter the buffer, and $\lambda$ is a scaling factor that adjusts the insertion probability based on the deviation of the actual proportion of old task data, $p_{\text{old}}$, from the target proportion, $w_{\text{buffer}}$. When $p_{\text{old}} > w_{\text{buffer}}$, the insertion probability increases, allowing more new task data to enter the buffer and reducing the dominance of old task data. Conversely, when $p_{\text{old}} < w_{\text{buffer}}$, the insertion probability decreases, prioritizing the retention of old task data. This dynamic adjustment mechanism ensures that the replay buffer is adaptively managed to suit the learning needs of different tasks and effectively leverage historical knowledge.

\textbf{Loss Function Design.} In terms of the loss function design, the total loss in SDW consists of two components: the policy optimization loss $\mathcal{L}_{\text{policy}}$ and the strategy consistency loss $\mathcal{L}_{\text{consistency}}$, expressed as:
$\mathcal{L}_{\text{total}} = \mathcal{L}_{\text{policy}} + \lambda_{\text{consistency}} \cdot \mathcal{L}_{\text{consistency}}.$
Here, $\mathcal{L}_{\text{policy}}$ focuses on optimizing the policy for the current task, enabling it to meet the objectives of the new task, while $\mathcal{L}_{\text{consistency}}$ constrains the current policy to maintain consistency with the historical policy, facilitating knowledge transfer. The dynamic adjustment of $\lambda_{\text{consistency}}$ is also derived from the similarity vector $\mathbf{S}$, reflecting the relationship between tasks and the need for knowledge transfer. 

\section{Experiments}
\subsection{Experimental Setup}
To evaluate the effectiveness of our proposed method, we conducted experiments on a series of continual learning problems. Specifically, we adopted a multi-round training framework wherein tasks in the environment were trained in multiple batches. In each batch, all tasks were trained sequentially, with the current model evaluated on all tasks at fixed intervals during training. This framework is widely used for evaluating lifelong learning methods, such as those in the Cora benchmark\cite{powers2022cora}.
We assessed the performance of the algorithm using the following metrics:
\subsubsection{Metrics}
To facilitate the definition of metrics, we first introduce the following notations:\\
- \( n \): The total number of tasks in the sequential learning setting.\\
- \( r_{i,j} \): The performance of the model on task \( i \) before and after training on task \( j \).\\
- \( r_{i,\text{all},\text{max}} \): The maximum performance of the model on task \( i \) across all evaluations.\\
- \( P \): The task performance metric.\\
- \( F \): The forgetting metric.\\
- \( T \): The knowledge transfer metric.

We now define the metrics used to evaluate the performance of different algorithms in sequential learning:
\begin{itemize}
    \item Task Performance (P)\\
    This metric evaluates the overall performance of the model across tasks:
    $$P = \frac{1}{n} \sum_{j=1}^{n} \left( \frac{1}{j} \sum_{i=1}^{j} r_{i,j} \right),$$ 
    \item Catastrophic Forgetting (F)\\
    This metric quantifies the amount of knowledge lost from previously learned tasks after learning subsequent tasks:
    $$
    F = \frac{1}{n-1} \sum_{j=2}^{n} \left( \frac{1}{j-1} \sum_{i=1}^{j-1} \frac{r_{i,j-1} - r_{i,j}}{|r_{i,\text{all},\text{max}}|} \right)\quad \text{s.t. } i < j,
    $$
    A positive value of $F$ indicates that the model’s performance on previous tasks has deteriorated after training on subsequent tasks.
    \item Knowledge Transfer (T)\\
    This metric measures how much prior knowledge contributes to learning new tasks:
    $$
    T = \frac{1}{n} \sum_{j=1}^{n} \left( \frac{1}{n-j} \sum_{i=j+1}^{n} \frac{r_{i,j} - r_{i,j-1}}{|r_{i,\text{all},\text{max}}|} \right) \quad \text{s.t. } i > j.
    $$
    A positive value of $T$ indicates that learning previous tasks improves the learning of subsequent tasks.
\end{itemize}
\subsection{Evaluation on the Minihack Environment}
Minihack is a sandbox environment created by \cite{samvelyan2021minihack}, based on the NetHack game. It uses the NLE interface to communicate with the game and provides a rich set of diverse task environments.
For our experiments, we selected 15 tasks from the Minihack environment, categorized into seven groups. Detailed task descriptions and the full task list are provided in the appendix.
We configured the training framework for two rounds, with each task trained for 1e7 steps per round, resulting in a total of 3e8 steps. To validate the generality of our approach, we conducted experiments using three LLMs—GPT-4o, GPT-3.5, and GLM4-9B—and compared our method against five lifelong reinforcement learning (LRL) methods. For all baseline algorithms, we use the default hyperparameter settings as provided in their official implementations unless otherwise stated.

\begin{table}
    \centering
    \scalebox{1}{
    \begin{tabular}{lrrr}
        \toprule
        \quad  & P $\uparrow$ & F $\downarrow$ & T $\uparrow$ \\
        \midrule
        EWC     & -0.216   & 0.175    & 0.217\\
        Clear   & -0.165   & 0.055       & -0.330\\
        Online-EWC& -0.289    & 0.266     & 0.177 \\
        PnC         & -0.286   & 0.075     & 0.150 \\
        SANE        & -0.162  & 0.354   & 0.097   \\
        \hline
        
        SDW+GPT4o   & \pmb{-0.151}  & \pmb{-0.505}  & \pmb{0.290}\\
        SDW+GPT3.5  & -0.160  & -0.096  & 0.194\\
        SDW+GLM4-9B & -0.152 & -0.264 & 0.134\\
        \bottomrule
    \end{tabular}
    }
    \caption{SDW performance comparison on Minihack.}
    \label{tab:table2}
\end{table}

\begin{figure}[htbp]
\centering 
\includegraphics[width=\linewidth]{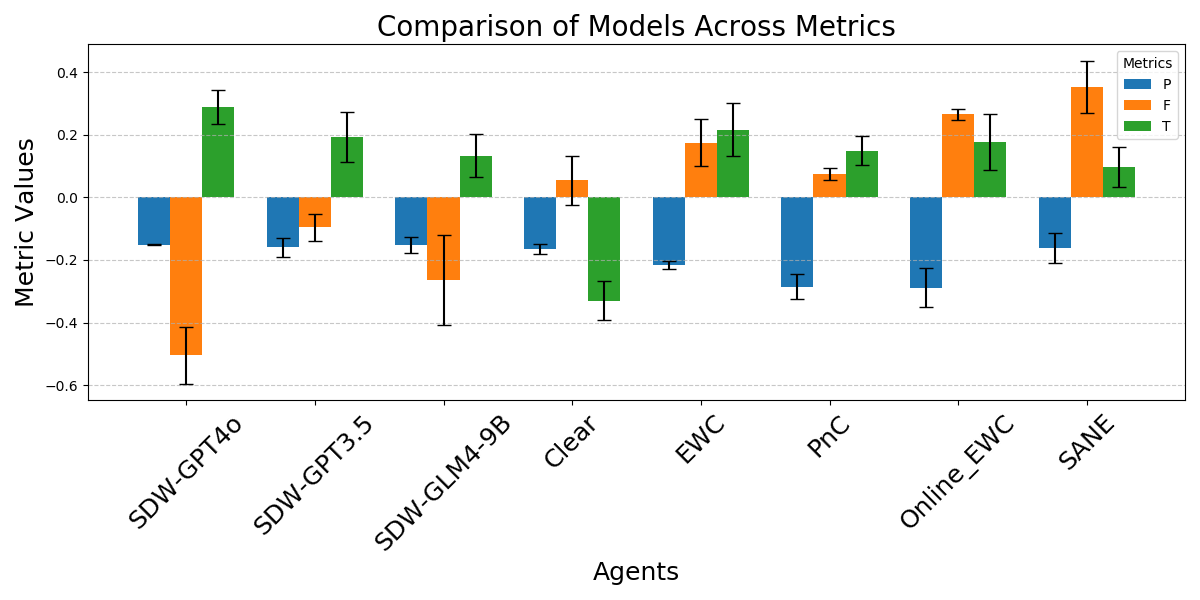}
\caption{SDW performance comparison on Minihack.} 
\label{fig_ex_minihack} 
\end{figure}
The experimental results in Table\ref{tab:table2} highlight the effectiveness of SDW in addressing lifelong learning challenges within the Minihack environment. When paired with LLMs such as GLM4-9B and GPT-4o, SDW demonstrates substantial performance improvements over traditional methods. Specifically, SDW+GPT-4o achieves the best overall task performance (-0.151) and excels in mitigating catastrophic forgetting (-0.505), underscoring its robustness in retaining knowledge across tasks. Additionally, it achieves the highest knowledge transfer score (0.290), showcasing its ability to generalize and adapt to new tasks effectively. Compared to baseline approaches such as EWC and SANE, SDW consistently outperforms across all metrics, demonstrating its strong potential for advancing lifelong learning frameworks. 
\subsection{Evaluation on the Atari Environment}
The Atari environment, based on the Arcade Learning Environment (ALE), provides a diverse set of classic games and tasks, serving as a standard benchmark for reinforcement learning methods.\cite{mnih2013playing} In our experiments, we selected three representative tasks: SpaceInvadersNoFrameskip-v4, BeamRiderNoFrameskip-v4, and MsPacmanNoFrameskip-v4. Each task was trained over three rounds, with 5 million steps per round, resulting in a total of 45 million training steps.

We evaluated three large-scale models—GPT-4o, GPT-3.5, and GLM4-9B—and compared their performance with the same lifelong reinforcement learning baselines, ensuring consistency across all experiments. For all baseline algorithms, we use the default hyperparameter settings as provided in their official implementations.

The experimental results in Table \ref{tab:table1_atari} demonstrate that SDW variants consistently outperform the baselines across all metrics. In task performance (P), SDW+GPT3.5 achieves the highest score (966.1), closely followed by SDW+GPT4o (965.46), while baselines EWC (425.72) and Clear (932.46) fall significantly behind. For catastrophic forgetting (F), SDW methods effectively mitigate forgetting with negative rates (-0.058 for SDW+GPT4o, -0.045 for SDW+GPT3.5), in contrast to the positive rates of EWC (0.113) and Clear (0.054). Most notably, SDW+GPT4o excels in knowledge transfer (T), achieving the highest score (0.124), surpassing its counterparts (SDW+GPT3.5: 0.069, SDW+GLM4-9B: 0.078) and baselines (EWC: -0.137, Clear: -0.024). These findings highlight the superior adaptability and scalability of SDW+GPT4o in continual learning scenarios, likely due to its enhanced reasoning and prior knowledge integration. However, further investigation is needed to understand the relatively lower transfer performance of SDW+GLM4-9B and explore the scalability of SDW methods in more diverse tasks.

\begin{table}
    \centering
    \begin{tabular}{lrrr}
        \toprule
        \quad  & P $\uparrow$ & F $\downarrow$ & T $\uparrow$ \\
        \midrule
        EWC       & 425.72   & 0.113      & -0.137        \\
        Clear     & 932.46     & 0.054     & -0.024         \\
        \hline
        SDW+GPT4o   & 965.46  & \pmb{-0.058} & \pmb{0.124}       \\
        SDW+GPT3.5   & \pmb{966.1}   & -0.045    & 0.069     \\
        SDW+GLM4-9B  & 933.61  & -0.011    & 0.078        \\
        \bottomrule
    \end{tabular}    
    \caption{SDW performance comparison on Atari.}
    \label{tab:table1_atari}
\end{table}

\begin{figure}[htbp]
\centering 
\includegraphics[width=0.6\linewidth]{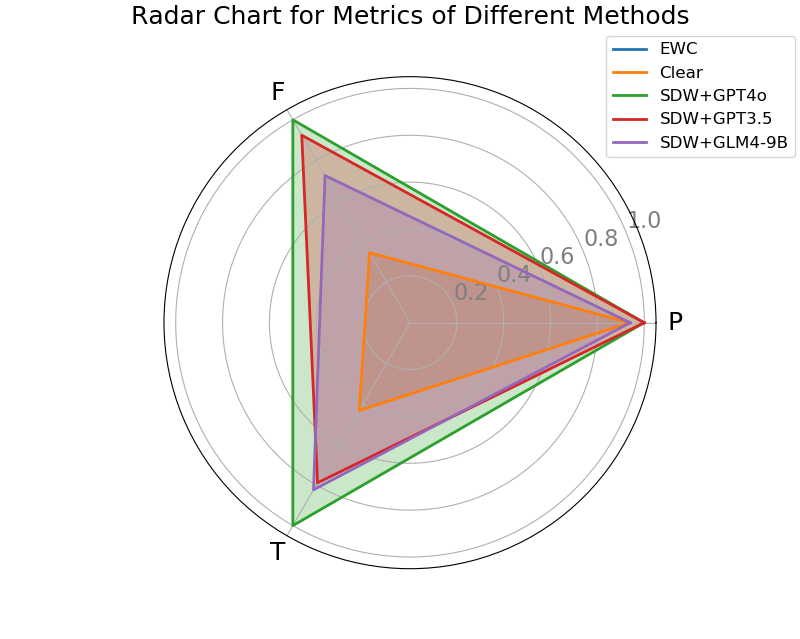}
\caption{SDW performance comparison on Atari.} 
\label{fig_ex_atari} 
\end{figure}

\subsection{Ablations}
To further investigate the effectiveness of our proposed SDW (Selective Dynamic Weighting) framework, we designed and tested several variants of SDW to isolate its contributions:
\begin{itemize}
    \item LLM-Based Loss and Buffer Module Generation(MG):
    By providing relevant code and task-specific information as input, large language models (LLMs) autonomously generate customized loss computation functions and buffer modules. 
    \item LLM-Based Parameter Assignment(PA):
    By leveraging task-specific information, task order, and reward functions as input, the large model dynamically configures parameters for loss computation and buffer management, enabling the agent to efficiently adapt and train for each specific task.
    \item SDW with loss function only(L)
    \item SDW with buffer function only(B)
\end{itemize}

Through a systematic comparison of these variants, we aim to delineate the specific contributions of key components within the SDW framework. The experiments are carried out in the MiniHack sandbox, adhering to the same setup detailed in Section 4.3. For this evaluation, ChatGPT-4o is utilized as the underlying LLM.

\begin{table}
    \centering
    \scalebox{1}{
    \begin{tabular}{lrrr}
        \toprule
        \quad  & P $\uparrow$ & F $\downarrow$ & T $\uparrow$ \\
        \midrule
        Clear   & -0.165    & 0.055       & -0.330\\
        \hline
        
        MG   & -0.384  & 0.580  & -0.325\\
        PA  & -0.165  & 0.075  & -0.939\\
        SDW(Only BUffer) & -0.164 & -0.260   & -0.066 \\
        SDW(Only Loss) & -0.166  & -0.341  & 0.259\\
        SDW(Buffer + Loss) & \pmb{-0.151} & \pmb{-0.505}  & \pmb{0.290}\\
        \bottomrule
    \end{tabular}
    }
    \caption{Ablation Study of SDW.}
    \label{tab:Ablation Study of SDW}
\end{table}
\begin{figure}[htbp]
\centering 
\includegraphics[width=\linewidth]{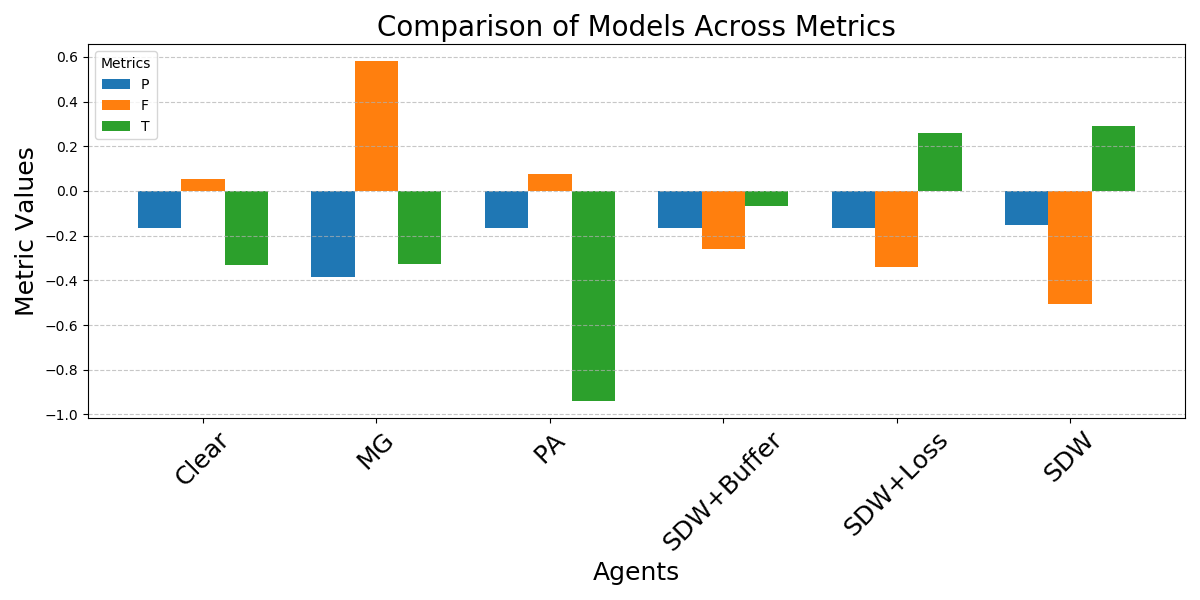}
\caption{Ablation Study of SDW.} 
\label{fig_ex_ablation} 
\end{figure}
The results in Table \ref{tab:Ablation Study of SDW} show that baseline methods (Clear, MG, and PA) exhibit poor balance across metrics. MG, due to the complete absence of chain-of-thought (CoT) reasoning, performs worst in catastrophic forgetting (Metric 2: 0.580 vs. Clear's 0.055), while PA shows the weakest knowledge transfer (Metric 3: -0.939 vs. Clear's -0.330). Adding the buffer component (SDW+Buffer) improves knowledge transfer (Metric 3: -0.066) but exacerbates forgetting (Metric 2: -0.260). Modifying the loss function (SDW+Loss) enhances knowledge transfer (Metric 3: 0.259) and reduces forgetting (Metric 2: -0.341) but fails to improve performance (Metric 1: -0.166). The complete SDW framework achieves the best balance, with the highest scores in knowledge transfer (Metric 3: 0.290) and performance (Metric 1: -0.151), and the lowest forgetting (Metric 2: -0.505). This suggests that the synergy between buffer and loss components is essential for optimizing results.

\section{Conclusion and Future Work}
In this paper, we proposed SDW, a novel framework for lifelong reinforcement learning that dynamically adjusts learning strategies based on task similarity. A key feature of SDW is its use of large pre-trained models to analyze task relationships, enabling a task similarity-driven weight computation mechanism that effectively balances knowledge retention and transfer. This design allows SDW to mitigate catastrophic forgetting while promoting efficient knowledge sharing across sequential tasks. Experimental results validate the effectiveness of SDW, demonstrating improved task performance, reduced forgetting, and enhanced knowledge transfer.

For future work, we aim to further leverage the capabilities of large-scale pre-trained models to replace manually designed components, such as reward function design and action-assisted generation. By automating these processes, SDW can become more generalized and adaptable to complex, open-ended lifelong reinforcement learning scenarios.

\clearpage
\newpage
\bibliographystyle{named}
\bibliography{main}

\clearpage
\appendix

\section{Insights into the Design and Mechanisms of SDW}

\subsection{Modular Design of SDW}

In the SDW framework, we avoid directly generating task weights \( w \), as this would require the LLM to search within a highly complex and high-dimensional function space \( \mathcal{F} \), mapping task descriptions to weight values. Instead, SDW adopts a structured, two-step design inspired by the Chain-of-Thought (CoT) paradigm. First, the LLM generates a task similarity function \( S \), which evaluates the relationships between tasks based on their descriptions, state distributions, and reward structures. This similarity function maps pairs of tasks to a similarity score \( S(T_{\text{new}}, T_{\text{old}}^{(i)}) \in [0, 1] \), providing a quantitative measure of task relationships in a relatively constrained function space \( \mathcal{S} \).

With the similarity scores as input, the LLM then generates a weight calculation function \( g \), which determines the training weights by considering both the similarity between the new task and old tasks as well as the relationships among old tasks. Formally, the weights are computed as 

\[
w = g\left(S(T_{\text{new}}, T_{\text{old}}^{(i)}), \{S(T_{\text{old}}^{(i)}, T_{\text{old}}^{(j)})\}_{j \neq i}\right).
\]

This second step operates in a smaller, structured subspace \( \mathcal{G} \), as it relies on a limited similarity matrix rather than the full task descriptions.

This two-step design offers several key advantages. First, it compresses the search space, as the complexity of generating \( S \) and \( g \) (i.e., \( \mathcal{S} + \mathcal{G} \)) is significantly lower than directly searching the full function space \( \mathcal{F} \). By effectively reducing the dimensionality of the problem, the LLM avoids blind exploration in a vast space and instead focuses on reasoning within smaller, interpretable subspaces. Second, the modular approach improves interpretability, since the task similarity function \( S \) provides an explicit representation of task relationships, making the reasoning process more understandable. Finally, this design enhances robustness, as isolating the complex weight computation step into a smaller subspace reduces the risk of instability or errors.

Mathematically, let \( \mathcal{F} \) denote the original function space for directly generating weights. By introducing the intermediate spaces \( \mathcal{S} \) for similarity functions and \( \mathcal{G} \) for weight calculation functions, the overall complexity is reduced such that 

\[
\mathcal{S} + \mathcal{G} \ll \mathcal{F}.
\]

The space \( \mathcal{S} \) is constrained by task feature dependencies, while \( \mathcal{G} \) depends only on structured similarity scores, both of which are significantly smaller than \( \mathcal{F} \). This reduction in complexity ensures a more efficient and reliable generative process while maintaining flexibility and interpretability.

\subsection{Dynamic Adjustment of Task Parameters}

It is important to note that \( w_{\text{buffer}} \) and \( \lambda_{\text{consistency}} \) are complementary mechanisms, addressing the influence of old tasks on the current task from two perspectives. \( w_{\text{buffer}} \) directly controls the proportion of old task data retained in the replay buffer, thereby influencing knowledge transfer. A higher \( w_{\text{buffer}} \) prioritizes the use of old task data, which is particularly effective when the new task is highly similar to previous tasks, allowing the model to efficiently reuse relevant knowledge. Conversely, when the new task is dissimilar, lowering \( w_{\text{buffer}} \) reduces interference from irrelevant or outdated data, ensuring the new task is learned independently and effectively.

On the other hand, \( \lambda_{\text{consistency}} \) adjusts the regularization that encourages consistency between the policies of old and new tasks. By aligning the new policy with established strategies, \( \lambda_{\text{consistency}} \) promotes smoother transitions and avoids abrupt changes that could degrade performance on old tasks. When task similarities are high, increasing \( \lambda_{\text{consistency}} \) ensures behavioral consistency and efficient knowledge reuse. However, when the new task requires greater independence, lowering \( \lambda_{\text{consistency}} \) allows the model to explore novel strategies without being overly constrained by past behaviors.

Both \( w_{\text{buffer}} \) and \( \lambda_{\text{consistency}} \) are dynamically adjusted based on the task similarity vector \( \mathbf{S} \). When \( \mathbf{S} \) indicates high similarity between the new and old tasks, SDW increases both \( w_{\text{buffer}} \) and \( \lambda_{\text{consistency}} \), ensuring greater data retention and behavioral consistency to maximize knowledge transfer. Conversely, when the new task requires more independence, SDW reduces these parameters to minimize interference, enabling the model to adapt more freely. This dynamic adjustment mechanism ensures that task similarities are leveraged when beneficial, while avoiding negative transfer or overfitting to old task strategies.

\section{Experimental Environment and Evaluation }
\subsection{Atari}
This section provides a concise overview of the three Atari environments used in our experiments: $SpaceInvadersNoFrameskip-v4$,\\$BeamRiderNoFrameskip-v4$, and \\$MsPacmanNoFrameskip-v4$. These environments, part of the Arcade Learning Environment (ALE), are instantiated using the Gym library and are designed with distinct objectives, action spaces, and observation formats. The following descriptions of the Atari Games are based on the official documentation in \url{https://www.gymlibrary.dev/environments/atari/complete_list/}

In $SpaceInvadersNoFrameskip-v4$, the player controls a laser cannon to destroy incoming space invaders before they reach Earth. Points are awarded for eliminating invaders, with higher scores for those in the back rows. The game ends when all lives are lost or the invaders reach the planet. In $BeamRiderNoFrameskip-v4$, the player pilots a spaceship, aiming to destroy enemy ships, dodge attacks, and avoid space debris, earning points for successfully eliminating enemies. Meanwhile, $MsPacmanNoFrameskip-v4$ tasks the player with navigating a maze to collect pellets while avoiding ghosts.

All three environments share a discrete action space of 18 actions, with game-specific reduced action sets available for more meaningful gameplay (e.g., movement, firing, or combined actions). Observations are provided as RGB images with dimensions (210, 160, 3) by default, but alternative observation formats, including 128-byte RAM representations and grayscale images, are also supported. Additionally, each environment allows for multiple modes and difficulty levels, offering flexible configurations for various experimental setups.

Version History: These environments have undergone several iterations. v0 represents the initial release, v4 removes action stickiness, and v5 reintroduces action stickiness while removing stochastic frameskipping. These updates improve the environments’ suitability for reinforcement learning research by ensuring a balance between complexity and reproducibility.

For further details regarding these environments and their configurations, please refer to the official documentation provided by the Arcade Learning Environment. The information summarized here is based on the documentation available at this source.

\subsection{Minihack}
This section provides an overview of the MiniHack environments used in our experiments. Each task is characterized by unique features and challenges, including procedurally generated maps, randomization, and varying levels of complexity. The following descriptions of the MiniHack environments are based on the official documentation provided by Mikayel Samvelyan et al. The full documentation is available at \url{https://github.com/facebookresearch/minihack}.
\subsubsection{Room:}
The Room tasks take place in a single square room, where the agent's objective is to navigate toward the staircase leading to the next level. There are multiple variants of this environment:
\begin{enumerate}
    \item \textbf{Room Sizes:} The room sizes are either 5x5 or 15x15.
    \begin{itemize}
        \item \texttt{MiniHack-Room-5x5-v0} and \\ \texttt{MiniHack-Room-15x15-v0}: Fixed starting and goal positions.
        \item \texttt{MiniHack-Room-Random-5x5-v0} and \\ \texttt{MiniHack-Room-Random-15x15-v0}: Randomized starting and goal positions.
    \end{itemize}
    
    \item \textbf{Complexity Additions:} Complexity increases through various factors:
    \begin{itemize}
        \item \textbf{Monsters:} Introduced in \\ \texttt{MiniHack-Room-Monster-5x5-v0} and \\ \texttt{MiniHack-Room-Monster-15x15-v0}.
        \item \textbf{Teleportation Traps:} Present in \\ \texttt{MiniHack-Room-Trap-5x5-v0} and \\ \texttt{MiniHack-Room-Trap-15x15-v0}.
        \item \textbf{Dark Rooms:} Where visibility is restricted to adjacent grid cells \\(\texttt{MiniHack-Room-Dark-5x5-v0} and \\ \texttt{MiniHack-Room-Dark-15x15-v0}).
        \item \textbf{Ultimate Challenge:} Combines all the above elements \\ (\texttt{MiniHack-Room-Ultimate-5x5-v0} and \\ \texttt{MiniHack-Room-Ultimate-15x15-v0}).
    \end{itemize}
    
    \item \textbf{Agent Capabilities:} The agent can attack monsters by moving into adjacent cells, but stepping on a lava tile results in instant death. In dark rooms, visibility is limited to adjacent cells.
    \item \textbf{Reward:} A reward of +1 is given upon reaching the goal.
\end{enumerate}
\subsubsection{Corridor Tasks}
The Corridor tasks challenge the agent to navigate procedurally generated rooms and corridors by using the \texttt{RANDOM\_CORRIDORS} command. The staircase leading to the next level is located in one of the rooms.
\begin{enumerate}
    \item \textbf{Randomized Layouts:} The positions and sizes of rooms and corridors are randomized for each episode.
    \item \textbf{Variants:} These tasks vary in complexity based on the number of rooms:
    \begin{itemize}
        \item \texttt{MiniHack-Corridor-R2-v0}: Two rooms.
        \item \texttt{MiniHack-Corridor-R3-v0}: Three rooms.
    \end{itemize}
    \item \textbf{Reward:} A reward of +1 is given upon reaching the goal.
\end{enumerate}

\subsubsection{KeyRoom Tasks}
The KeyRoom tasks require the agent to retrieve a key, navigate to a locked door, and unlock it to reach the staircase.
\begin{enumerate}
    \item \textbf{Task Features:}
    \begin{itemize}
        \item The key, door, and staircase locations are randomized.
        \item The agent's action space includes standard movement, pickup, and apply actions.
    \end{itemize}
    \item \textbf{Variants:}
    \begin{itemize}
        \item Room sizes include 5x5 (\texttt{MiniHack-KeyRoom-S5-v0}) and 15x15 (\texttt{MiniHack-KeyRoom-S15-v0}).
        \item Dark versions\\
        (\texttt{MiniHack-KeyRoom-Dark-S5-v0} and \texttt{MiniHack-KeyRoom-Dark-S15-v0}) restrict visibility to adjacent cells, making the key invisible unless nearby.
    \end{itemize}
    \item \textbf{Reward:} A reward of +1 is given upon reaching the goal.
\end{enumerate}

\subsubsection{River Tasks}
The River tasks require the agent to cross a river by pushing boulders into the water to create a path.
\begin{enumerate}
    \item \textbf{Task Features:}
    \begin{itemize}
        \item Boulders pushed into water create walkable tiles.
        \item The agent must avoid pushing boulders into lava, which results in failure.
    \end{itemize}
    \item \textbf{Variants:}
    \begin{itemize}
        \item Narrow river crossing (\texttt{MiniHack-River-Narrow-v0}).
        \item Addition of monsters (\texttt{MiniHack-River-Monster-v0}).
        \item Lava hazards (\texttt{MiniHack-River-Lava-v0}).
    \end{itemize}
    \item \textbf{Reward:} A reward of +1 is given upon reaching the goal.
\end{enumerate}

\subsubsection{HideNSeek Tasks}
The HideNSeek tasks place the agent in a procedurally generated room with trees, clouds, and a powerful monster.
\begin{enumerate}
    \item \textbf{Task Features:}
    \begin{itemize}
        \item Trees block movement and line of sight.
        \item Clouds obscure visibility but allow movement.
        \item The agent must avoid detection by the monster while navigating to the goal.
    \end{itemize}
    \item \textbf{Variants:}
    \begin{itemize}
        \item Standard version (\verb|MiniHack-HideNSeek-v0|).
        \item Lava hazards (\texttt{MiniHack-HideNSeek-Lava-v0}).
    \end{itemize}
    \item \textbf{Reward:} A reward of +1 is given upon reaching the goal.
\end{enumerate}

\subsubsection{CorridorBattle Tasks}
The CorridorBattle tasks challenge the agent to defeat a horde of monsters by strategically using narrow corridors.
\begin{enumerate}
    \item \textbf{Task Features:}
    \begin{itemize}
        \item Fighting in corridors allows the agent to engage monsters one at a time, minimizing damage.
        \item Dark versions \\
        (\texttt{MiniHack-CorridorBattle-Dark-v0}) require the agent to remember the number of defeated monsters to plan subsequent actions.
    \end{itemize}
    \item \textbf{Reward:} A reward of +1 is given upon reaching the goal.
\end{enumerate}

\section{Prompts and Code Generated by LLMs}
\subsection{Prompts}
This section provides the prompts we used in SDW experiments.
\begin{itemize}
    \item This prompt is used for task similarity function generation:\\ 
    \texttt{There is now a training environment containing n tasks, and the model will sequentially train on these n tasks. After completing the training of each task, the current model will be tested on all tasks. Based on the test results for all tasks, the model's catastrophic forgetting, knowledge transfer, and task performance will be calculated. Now, I hope you can design a method to compute task relevance based on the task information and the order of the tasks I provide below (Task Information).}\\(The task information consists of all the details provided in Section B.1 or Section B.2.) 
    \item This prompt is used for weight computation function(loss part) generation:\\ \texttt{Now, within this training framework (a training environment with n tasks, where the model sequentially trains on these n tasks. After completing the training of each task, the current model is tested on all tasks. Based on the test results for all tasks, the model's catastrophic forgetting, knowledge transfer, and task performance are calculated), you are required to design a function that calculates the parameter values based on the task similarity vector.The function to compute the loss is as follows: (relevant algorithm code).There are two parameters in this function: value cloning cost and policy cloning cost. This loss calculation method can prevent catastrophic forgetting during multi-task training but limits the model's exploration and training on individual tasks.Please design a function module that calculates the parameter values based on the task similarity. }
    \item This prompt is used for weight computation function(buffer part) generation:\\ \texttt{There is now an additional parameter: batch replay ratio, which represents the proportion of past task experiences in the buffer. I hope you can design a calculation method to determine the value of this parameter based on the task relevance vector.Please provide a calculation function where the input is the similarity vector, and the output is the parameter value. }

\end{itemize}

\subsection{Code}
This section provides the code generated by different LLMs using prompt in Section B.1.
\begin{itemize}
    \item GPT-4o
    \begin{lstlisting}[language=Python, caption={Task similarity function generated by GPT-4o}]
def compute_task_similarity_vector(self, agent_output_1, env_output_1, agent_output_2, env_output_2):
        frame_1 = env_output_1['frame'].float().flatten().numpy()
        frame_2 = env_output_2['frame'].float().flatten().numpy()
        js_distance = jensenshannon(frame_1, frame_2)
        state_similarity = 1 - js_distance  
        logits_1 = agent_output_1['policy_logits']
        logits_2 = agent_output_2['policy_logits']
        probs_1 = F.softmax(logits_1, dim=-1).squeeze().numpy()
        probs_2 = F.softmax(logits_2, dim=-1).squeeze().numpy()
        js_distance_policy = jensenshannon(probs_1, probs_2)
        policy_similarity = 1 - js_distance_policy 
        baseline_1 = agent_output_1['baseline'].item()
        baseline_2 = agent_output_2['baseline'].item()
        baseline_similarity = 1 - abs(baseline_1 - baseline_2)
        similarity_vector = np.array([state_similarity, policy_similarity, baseline_similarity])
        return similarity_vector
    \end{lstlisting}
    \begin{lstlisting}[language=Python, caption={Weight computation function generated by GPT-4o: Part i}]
def compute_cloning_costs(self, similarity_vector, max_policy_cost=0.01,    max_value_cost=0.005, min_cost=-0.0001):
    state_sim = similarity_vector[0]
    policy_sim = similarity_vector[1]
    policy_cloning_loss_cost = max_policy_cost *(0.8 * policy_sim + 0.2 * state_sim)
    policy_cloning_loss_cost s= max(policy_cloning_loss, min_cost) 
    value_cloning_loss_cost = max_value_cost *(1- (0.9 * value_sim + 0.1 * state_sim))
    value_cloning_loss_cost = max(value_cloning_loss, min_cost)  
    return policy_cloning_loss_cost, value_cloning_loss_cost
    \end{lstlisting}
    \begin{lstlisting}[language=Python, caption={Weight computation function generated by GPT-4o: Part ii}]
def compute_batch_replay_ratio(self, similarity_vector, base_ratio=0.8, smooth_factor=2):
    state_sim = similarity_vector[0]
    policy_sim = similarity_vector[1]
    value_sim = similarity_vector[2]
    similarity = 0.4 * state_sim + 0.4 * policy_sim + 0.2 * value_sim 
    batch_replay_ratio = base_ratio + (1 - base_ratio) * (1 - similarity**smooth_factor)
    return batch_replay_ratio
\end{lstlisting}
    \item GPT-3.5
    \begin{lstlisting}[language=Python, caption={Task similarity function generated by GPT-3.5}]
def compute_task_similarity_vector(self, agent_output1, env_output1, agent_output2, env_output2):
    policy_logits1 = agent_output1['policy_logits']
    baseline1 = agent_output1['baseline']
    frame1 = env_output1['frame']
    episode_return1 = env_output1['episode_return']
    policy_logits2 = agent_output2['policy_logits']
    baseline2 = agent_output2['baseline']
    frame2 = env_output2['frame']
    episode_return2 = env_output2['episode_return']
    feature_vector1 = torch.cat((policy_logits1.view(-1), baseline1.view(-1), frame1.view(-1), episode_return1.view(-1)))
    feature_vector2 = torch.cat((policy_logits2.view(-1), baseline2.view(-1), frame2.view(-1), episode_return2.view(-1)))
    similarity_score = self.calculate_cosine_similarity(feature_vector1, feature_vector2)
    return similarity_score
    \end{lstlisting}
    \begin{lstlisting}[language=Python, caption={Weight computation function generated by GPT-3.5: Part i}]
def compute_cloning_costs(self, similarity, max_policy_cost=0.01, max_value_cost=0.005, min_cost=-0.0001):
    if similarity > 0.8:
        value_cloning_loss_cost = 0.0
        policy_cloning_loss_cost = 0.0
    elif similarity > 0.6:
        value_cloning_loss_cost = 0.0
        policy_cloning_loss_cost = 0.01
    elif similarity > 0.4:
        value_cloning_loss_cost = 0.01
        policy_cloning_loss_cost = 0.0
    else:
        value_cloning_loss_cost = 0.01
        policy_cloning_loss_cost = 0.01
    return value_cloning_loss_cost, policy_cloning_loss_cost
    \end{lstlisting}
    \begin{lstlisting}[language=Python, caption={Weight computation function generated by GPT-3.5: Part ii}]
def compute_batch_replay_ratio(self, similarity_vec, initial_value=0.5, max_value=1.0):
    initial_ratio = 0.5
    max_ratio = 1.0
    similarity = np.average(similarity_vec)
    if similarity >= 0.8:
        return max_ratio
    else:
        return initial_ratio + (max_ratio - initial_ratio) * similarity
\end{lstlisting}
    \item GLM4-9B
    \begin{lstlisting}[language=Python, caption={Task similarity function generated by GLM4-9B}]
def compute_task_similarity_vector(self, agent_output_1, env_output_1, agent_output_2, env_output_2):
        policy_logits1 = agent_output_1['policy_logits']
        policy_logits2 = agent_output_2['policy_logits']
        action1 = agent_output_1['action']
        action2 = agent_output_2['action']
        baseline1 = agent_output_1['baseline']
        baseline2 = agent_output_2['baseline']
        policy_similarity = self.cosine_similarity_custom(policy_logits1, policy_logits2)
        action_similarity = self.jaccard_similarity_custom(set(action1), set(action2))
        baseline_similarity = 1 - np.abs(baseline1 - baseline2) / np.max([np.abs(baseline1), np.abs(baseline2)])
        similarity_vector = np.array([policy_similarity, action_similarity, baseline_similarity])
        return similarity_vector
    \end{lstlisting}
    \begin{lstlisting}[language=Python, caption={Weight computation function generated by GLM4-9B: Part i}]
def compute_cloning_costs(task_similarity_vector, default_policy_cost=0.01, default_value_cost=0.005, min_cost=0.001):
    similarity_weights = 1 / (1 + np.exp(-task_similarity_vector))
    average_weight = np.mean(similarity_weights)
    value_cloning_cost = default_value_cost * average_weight
    policy_cloning_cost = default_policy_cost * average_weight
    return value_cloning_cost, policy_cloning_cost
    \end{lstlisting}
    \begin{lstlisting}[language=Python, caption={Weight computation function generated by GLM4-9B: Part ii}]
def compute_batch_replay_ratio(self, similarity, initial_value=0.5, max_value=1.0):
    normalized_similarity = np.clip(similarity, 0, 1)
    exponent = 5
    batch_replay_ratio = initial_value + (max_value - initial_value) * np.log(normalized_similarity) / np.log(1 / (1 - initial_value))
    batch_replay_ratio = np.clip(batch_replay_ratio, initial_value, max_value)
    return batch_replay_ratio
\end{lstlisting}
\end{itemize}

\end{document}